\documentclass[10pt,conference]{IEEEtran}
\IEEEoverridecommandlockouts
\usepackage{cite}

\usepackage{amsmath}
\usepackage{amssymb}
\usepackage{enumerate}
\usepackage{graphicx}
\usepackage{textcomp}
\usepackage{xcolor}
\usepackage[super]{nth}

\usepackage{subcaption}    

\usepackage{balance}
\usepackage{hyperref}
\usepackage{authblk}
\usepackage{booktabs}
\usepackage{graphicx}
\usepackage{algpseudocode}
\usepackage{textcomp}
\usepackage{xcolor}
\usepackage{url}
\usepackage{graphicx}
\usepackage{subcaption}
\usepackage{verbatim}
\hypersetup{bookmarks=false}
\usepackage{algorithm}
\usepackage{algorithmicx}
\newlength\myindent
\def\BibTeX{{\rm B\kern-.05em{\sc i\kern-.025em b}\kern-.08em
    T\kern-.1667em\lower.7ex\hbox{E}\kern-.125emX}}
\begin{document}

\makeatletter
\newcommand{\linebreakand}{%
  \end{@IEEEauthorhalign}
  \hfill\mbox{}\par
  \mbox{}\hfill\begin{@IEEEauthorhalign}
}
\makeatother

\newcommand{\Secref}[1]{Sec.~\ref{#1}}


\title{Drift-Based Dataset Stability Benchmark 
\thanks{This research was funded by the Ministry of Interior of the Czech Republic, grant No. VJ02010024: Flow-Based Encrypted Traffic Analysis and also by the Grant Agency of the CTU in Prague, grant No. SGS23/207/OHK3/3T/18 funded by the MEYS of the Czech Republic.}}

\author[1]{Dominik Soukup}
\author[1]{Richard Plny}
\author[1]{Daniel Vašata}
\author[2]{Tomáš Čejka}
\affil[1]{Czech Technical University in Prague, Prague, Czech Republic}

\affil[2]{CESNET, a.l.e., Prague, Czech Republic \authorcr {\tt \{soukudom, plnyrich, vasatdan\}@fit.cvut.cz, cejkat@cesnet.cz}}

\maketitle

\begin{abstract}
    Machine learning (ML) represents an efficient and popular approach for network traffic classification. However, network traffic classification is a challenging domain, and trained models may degrade soon after deployment due to the obsolete datasets and quick evolution of computer networks as new or updated protocols appear. Moreover, significant change in the behavior of a traffic type (and, therefore, the underlying features representing the traffic) can produce a large and sudden performance drop of the deployed model, known as a data or concept drift. In most cases, complete retraining is performed, often without further investigation of root causes, as good dataset quality is assumed. However, this is not always the case and further investigation must be performed. This paper proposes a novel methodology to evaluate the stability of datasets and a benchmark workflow that can be used to compare datasets.
    The proposed framework is based on a concept drift detection method that also uses ML feature weights to boost the detection performance. The benefits of this work are demonstrated on CESNET-TLS-Year22 dataset. We provide the initial dataset stability benchmark that is used to describe dataset stability and weak points to identify the next steps for optimization. Lastly, using the proposed benchmarking methodology, we show the optimization impact on the created dataset variants.

\end{abstract}


\section{Introduction}

    Network traffic monitoring provides an essential insight for maintaining services and security, and network traffic classification is its integral part. The security features, such as the \textit{Encrypted Server Name Indication (ESNI)} for domain names encryption, forces the development of new means of network monitoring and analysis. Researchers have thus been focusing on indirect network inspection, identification of metadata, investigation of packet series and their temporal statistical properties, and analysis that does not rely on decryption.
    For example,  \cite{aceto2021distiller,luxemburk2023fine,malekghaini2023deep,dai2023glads,10199052,akbari2021look} target on temporal network characteristics in Sequence of Packet Lengths and Times (SPLT) method, Koumar et al. \cite{koumar2024nettisa} suggested extended IP flow NetTiSA for network traffic classification and Plny et al. \cite{plny2023decrypto} present a detection method based on multiple weak indicators.

    Research of network traffic classification drives the need to deploy ML classifiers and maintain them continually. In real-world environments, the collected data often tends to change over time, e.g. reflecting state of CDNs, protocol modifications and updates, used encryption methods, and more. This phenomenon is referred as concept or data drift, and network traffic inspection should ideally maintain its accuracy irrespective of such externalities. A possible solution is the concept of Active Learning (AL) that updates underlying datasets used for periodic retraining. However, due to the dynamicity of investigated network services, precise tuning of AL parameters is notoriously demanding.
    For this optimization we need to know details about the used dataset so that we can identify weak points and improve the dataset properly. In the community, this is typically done by dataset metrics~\cite{dataset-metric}, however, they are not always feasible for long-term dataset evaluation and do not support optimization. 
    To address these challenges, it is crucial to understand the behavior and importance of selected features in time, and the relationships between different features and classes for the targeted use case. Moreover, it is important to have unified workflow to evaluate dataset optimization results.  

    This paper presents an universal workflow for dataset stability benchmarking based on concept drift detection. The goal is to evaluate stability of a dataset with possibility to identify changes in the dataset and investigate features and classes that caused these drifts. Results of this benchmarking workflow can be used for offline dataset evaluation (to identify if it is suitable for the task, e.g., traffic classification), dataset optimization (suggest how to split dataset or identify weak spots for further investigation), or continuous dataset monitoring that triggers AL dataset update. 
    ML-model retraining is a computationally expensive task and it is important to know which part of the dataset is impacted to run retraining strategy efficiently. As part of the newly proposed benchmarking workflow, we introduce a novel drift detection method called MFWDD (Model-based Feature Weight Drift Detection) for ML-models based on feature weights. The method can detect model obsoletion in supervised and unsupervised cases to support real-world deployment scenarios. 

    The main contributions of this paper can be summarized as follows:
     \begin{enumerate}[I)]
         \item We introduced universal dataset benchmark workflow that can evaluate dataset stability.
         \item We proposed model-based feature weighted drift detection method for supervised and unsupervised scenarios. In supervised scenarios, the technique brings descriptive analysis of behavioral changes in input features and classes which can be used for dataset optimization.
         \item We applied the methodology to large and long-term dataset CESNET-TLS-Year22 and demonstrated results on specific dataset variants to provide full understanding.
         \item We made the dataset benchmarking tool available for the community at Github repository\footnote{https://github.com/FETA-Project/MFWDD} and released public repository with evaluation of other datasets\footnote{https://dataset-catalog.liberouter.org/} and dataset metrics.
     \end{enumerate}
  
    This paper is divided as follows: Section~\ref{sec:related_works} summarizes the related work on network traffic dataset drift and benchmark analysis. Section~\ref{sec:mfwdd} proposes a novel approach for detecting data drifts. Section~\ref{sec:benchamrk} introduces a universal benchmarking workflow for dataset stability evaluation. Section~\ref{sec:analysis} 
    contains two case studies with detailed experiments to show the benefits of the proposed method and our findings.
 Section~\ref{sec:discussion} discusses the outcomes of our work and recommends best practices for the domain of network traffic classification. Section~\ref{sec:future} presents future work based on the achieved results. Lastly, Section~\ref{sec:conclusion} concludes the paper and addresses research directions for future work.

\section{Related works} \label{sec:related_works}

In this section, we review related works. The section is divided into I) Drifts, and II) Dataset Benchmarking, to increase its readability.

\subsection{Drifts}

Data drift and concept drift were surveyed in~\cite{drift_survey,datadrift}. Data drift occurs when the statistical properties of the input data change, leading to a potential degradation in model performance because the model was trained on a different data distribution. Concept drift, on the other hand, refers to changes in the target variable's definition or the relationship between input features and the target variable. Both types of drift pose significant challenges for predictive models in dynamic environments requiring continuous monitoring and model maintenance ensuring their effectiveness and relevance.

There are several existing methods to detect a concept drift. A common example is \textit{error rate-based drift detection}. These algorithms look at the error rate of a particular classifier. If the change of error rates over time is statistically significant, a drift alarm is raised. Another common method is \textit{distribution-based drift detection}. In this case, algorithms use distance metrics to detect changes between the original and new training data. Korycki et al.~\cite{imbalanced-drift} point out detection limits in network traffic environment and introduced a drift detector called RBM-IM. It is based on a Restricted Boltzmann Machine, and it is designed for imbalanced and multi-class streams, which are a common classification task in the field.

To address the sensitivity limits of dataset drift detection methods, weighting or sampling is used to adjust for the differences in the data distributions. Soukup et al.\cite{soukup2021towards} introduced the idea of dataset quality evaluation, which is affected by the metadata used by the ML classifier. This work focuses on ML-assisted network traffic classification problems and continuous model maintenance compensating for concept drifts. We attempted to leverage feature importance as weights to calculate drift severities, which yielded promising results. There are multiple methods to integrate various feature distributions into the final decision regarding the presence of a drift and its severity. A popular library in this field is Evidently~\cite{evidently_comparison}, which utilizes the two-sample Kolmogorov-Smirnov (KS) test or the Wasserstein distance. Based on the results in \cite{evidently_comparison}, the Wasserstein distance is recommended for datasets with more than a thousand data instances. Otherwise, the KS test gives better results. Other statistical tests include Maximum Mean Discrepancy (MMD) two-sample test~\cite{drift_survey2} and Page-Hinkley test (PHT)~\cite{page1954continuous}. The PHT test was originally addressing production quality problems, and inspection schemes optimization. While investigating network intrusion detection systems, Andresini et al.~\cite{andresini2021network} successfully used PHT as a drift estimator in the Str-MINDFUL algorithm. Their experimental setting, unlike ours, aimed at labeling malicious and benign network traffic from IP flow streams into approximately 10 classes. In our scenario, we consider much more complex labeling and service detection, and thus, their proposed retraining could be less efficient.
We follow up on the existing state of the art and evaluate main metrics within the networking domain with the proposed feature weights. The introduced software tool is configurable and allows to change the statistical test if needed.

\subsection{Dataset Benchmarking}

Machine learning performance is highly affected by quality of the underlying datasets, known as the famous ''garbage in, garbage out'' principle. Missing data, used preprocessing and cleaning methods, bias, and more affect the quality of the dataset~\cite{gong2023survey}. However, even well-known and widely used network traffic datasets suffer from certain issues. Lavin et al.~\cite{lanvin-errors-CICIDS2017} identified several issues in the CIC-IDS-2017 dataset. Moreover, previous work also identified other errors in data collection and labeling processes. In addition, inconsistencies were also found by Aceto et al.~\cite{aceto2021distiller} in another widely used CIC-VPN-nonVPN-2016 dataset.


In addition, network traffic datasets quickly become obsolete, as a computer network is highly dynamic environment and a dataset only captures the state of the network during dataset collection period~\cite{pesek2023alf}. Guerra et al.~\cite{guerra2022datasets} focus on labeling methods and identify significant problems affecting dataset quality, such as that quality of automatic labeling cannot be guaranteed. Moreover, there is a trade-off between the quality of resulting labeling and the size of the dataset. As a result, network traffic datasets might be poorly labeled, as they have thousands and millions of samples.

Benchmarking is a known concept that is especially done for ML models. For these types of benchmarks, users select a list of well-known datasets and compare achieved results~\cite{wils, rituals}. In this work, we focus on the dataset level benchmarking. 
Benchmarking methods aim to assess dataset quality and identify potential issues, such as improper labeling. Several studies~\cite{hamid2018benchmark,guerra2022datasets,layeghy2024benchmarking} evaluated chosen TC datasets, however, no unified dataset benchmark framework have been proposed to our knowledge. We argue that a reproducible workflow for dataset quality evaluation and assessment is needed, to obtain results for each dataset that can be compared to each other.

\section{MFWDD: Model-based Feature Weight Drift Detection Methodology} \label{sec:mfwdd}
    In this section, we present details of the proposed method, MFWDD, for feature weighted detection of concept drifts, which is an essential part of the dataset stability benchmark workflow. To detect a distribution drift at a given time, the drift detector compares samples from the initial and current distributions. The detector performs statistical tests and evaluates whether pairs of samples originate from different (or the same) distributions. Upon detecting a drift and its measure, a trigger prompting to retrain the classification model is raised. The measure should quantify dissimilarity between the two distributions and reflect the performance impact of network traffic classification. It will be referred to as drift severity and denoted by $s$.

    Our experiments with KS test and Wasserstein distance confirm the same best practice. Although the KS test was more sensitive and often identified more features as drifted for datasets with more than approximately a thousand samples, it provided more precise results for smaller data collections. For class (supervised) level detection, we used the KS test with a significance level threshold of $\alpha = 5\% $, since the number of samples per class per day was below one thousand. In global (unsupervised) scenarios, the normalized Wasserstein distance with a threshold of $ \alpha = 5\% $ was used. Both the KS test and normalized Wasserstein distance use a single variable vector and, when applied on the $i$-th feature, they yield severity $s_i$. 
    
    During our investigation, we also considered the MMD test, which operates with a multivariate distribution of more features and theoretically can detect more complex dependencies. Despite experimenting with various kernel settings and sensitivity levels, MMD did not prove beneficial for our classification tasks. Moreover, analyzing drifts for specific features would be more complex with MMD. Thus, we decided to build our method on the Wasserstein distance and the KS test. Even though the selected threshold can be optimized for each dataset and ML classifier, we focused on the identification of suitable default values. The goal of this work is to create dataset stability benchmark where we want to have the same input thresholds.

    Considering a classification task with $n$ features and given a statistical measure, i.e. two-sample Kolmogorov-Smirnov (KS) test or the Wasserstein distance, we may calculate individual drift measures $s_1, ..., s_n$ for each feature separately and then investigate their possible combinations given by a mapping $S:~s_1, ..., s_n~\mapsto~\mathbb{R}_+$ called \textit{severity} or \textit{drift severity}. A straightforward enhancement may incorporate feature weights as an additional measure for the drift severity evaluation. The drift severity mapping $S$ is then of the form $S: s_1, ..., s_n, w_1, ..., w_n \mapsto \mathbb{R}_+.$
    
    Current drift detection methods often neglect the significance of individual features in affected models, and they are sensitive to reporting false positives since they neglect the target dataset context and externalities. This is particularly challenging when feature vectors contain hundreds of features, some of which provide minimal information.
    
    To address this issue, statistical tests are conducted independently for each feature distribution, and their drift severities are stored for further analysis, for example: drift rates, extraction of features exhibiting severe drift, features with recurring drifts etc. The total drift strength $S$ is calculated as a weighted arithmetic mean of individual severity measures $s_i$, where the feature importances supplied by the classifier serve as the weights $w_i$. In our experiments, we used a normalization convention $ \sum_{i=1}^n w_i = 1$, where $w_i$ are model feature importances.   
    We propose to use this normalization convention for benchmarking compatibility; other choices require a drift detection threshold renormalization. 
    

    Thus, the overall drift severity is computed as by the equation $S = \sum_{i=1}^n w_i s_i $ where $ n $ is the number of features, $s_i$ $i$-th drift measure and $w_i$ $i^{th}$ drift weight. As a result, we arrive at drift strength representation incorporating intrinsic dataset statistical properties. We may state that a drift is detected when the total severity $S \ge s_{t}$, where $s_{t}$ is a threshold, e.g. $s_{t} = 0.5$. Comparison of results with and without weights using feature importances is described in the Sec.~\ref{sec:analysis}.
    
    This method can be applied in both supervised and unsupervised scenarios. In unsupervised scenarios, it detects distribution changes that likely decrease model performance globally. Consequently, we can initiate an annotation process, employ strategies for updating the dataset and retrain the model. In supervised scenarios, we identify performance decrease using metrics such as accuracy or F1 score. However, the proposed drift analysis provides other useful data: which features exhibit drift, how severe the drifts are, are they gradual, recurring etc. Such insight helps us to react accordingly, for instance, by excluding features that frequently cause drifts.

      

\section{Dataset stability benchmark methodology}\label{sec:benchamrk}
The introduced MFWDD method is important for quickly comparing two dataset time windows and identifying possible drifts. Typically, one window represents the current dataset, and the second window contains data from the current day. However, it is important to have a unified workflow that allows us to evaluate long-term datasets and provide benchmark results that will allow us to track changes in detail. To achieve this goal, we defined the procedure shown in Alg.~\ref{alg:MFWDD_benchmark}.
The algorithm consists of three stages: \textit{Train}, \textit{Configure}, and \textit{Evaluate}. In the \textit{Training} stage, we take the training part of the dataset (by default, one week) and two ML models. One for retraining workflow and the second for reference workflow (no retraining). In the \textit{Configuration} stage, we initialize the MFWDD module with feature importances as weights and statistical tests, described in~Sec.~\ref{sec:mfwdd}. 
The essence of the \textit{Evaluation} stage is that for each window in the evaluated dataset, firstly, a prediction by machine learning is made for that window, by both models. Then, the prediction and ground truth data are compared (line 10) and drift detection is performed. If the drift is detected, the algorithm replaces the oldest samples with the ones from the currently processed window and retrains the retraining ML model (lines 12 and 13). The referenced ML model is not retrained. Lastly, we store logs for further visualization and analysis. 

\algrenewcommand\algorithmicforall{\textbf{for each}}

\begin{algorithm}
\caption{Dataset stability benchmark workflow}
\label{alg:MFWDD_benchmark}
\begin{algorithmic}[1]
\State \textbf{Stage Train:} Train ML models to run evaluation workflows and get feature importance for weights
\State ML\_ref $\gets$ ML(training window)
\State ML\_retrain $\gets$ ML(training window)
\State ref\_window $\gets$ training window
\State \textbf{Stage Configure:} Initialize MFWDD detection module 
\State MFWDD $\gets$ MFWDD(config\_params)
\State \textbf{Stage Evaluate:} Dataset evaluation loop
\ForAll{\texttt{window} $\in$ dataset}
\State prediction $\gets$ ML(window)
\State drift $\gets$ MFWDD(prediction, ref\_window)
\If{drift detected}
    \State replace oldest samples from \texttt{ref\_window} with  \texttt{window}
    \State retrain ML\_retrain
\Else
    \State continue
\EndIf
\EndFor
\State Store logs for benchmark processing and visualization
\end{algorithmic}
\end{algorithm}

The defined workflow is very flexible in terms of configuration. We can set windows using dataframe (e.g. one day) or using amount of samples (e.g. 10000). Therefore, we keep the initial configuration part of benchmark logs for easy comparison. 

We also define visualization schema for easy interpretations and comparison of received logs. For each category, we consider results without retraining and results with retraining to easily see the impact.
Visualization is done in the following categories:
    \subsection*{Global Results}
    This section contains drift detection results for the complete input dataset (global level). We provide summary of total amount of drifts, drifted features and F1 score. These values are supplemented with description of input configuration (statistical tests and thresholds) and visualization of the main evaluation loop with all windows described in Alg.~\ref{alg:MFWDD_benchmark}. 
    \subsection*{Per Class Analysis}
    If the input dataset contains labels, we show focused view for each class. Since the list of classes can be large, we by default limit the visualisation to top five most drifted classes. Provided output contains drift strength and F1 score for each class correlated with global drift detections for all windows. Moreover, we provide list of the most correlated classes based on the calculated drift strength and F1 score. With these results, we can identify problematic classes overtime and decide how to split the dataset to get more stable results and more efficient retraining.  
    \subsection*{Per Feature Analysis} 
    This section provides visualisation of drift strength for each feature separately. Since the list of features can be long, we by default show only top five features with the highest drift strength. The visualisation is correlated with detected global drift to track when the retraining model was updated. Based on this output, we can analyze behavior of each feature or group of features and assess the suitability of features set with the selected ML model.  
    

Demonstration of the proposed workflow together with interpretation is described in the next section with different case studies. Full reports, even with detailed logs, are available in the public repository\footnote{https://dataset-catalog.liberouter.org/}.


\section{Experiments and Analysis} \label{sec:analysis}
    The aim of this section is to demonstrate the performance of the introduced workflow with MFWDD detection. Since we aim to find out how the distribution of various features evolves in time, we chose a one-year-long CESNET-TLS-Year22~\cite{hynek_2024_10608607} dataset, accessed through CESNET-DataZoo library~\cite{luxemburk2023datazoo}. It contains network traffic flows captured on backbone lines of ISP-level network and contains more than a hundred network traffic classes. The dataset size allows us to flexibly arrange diverse experimental settings, perform an offline evaluation of drift behavior concerning various research topics, and test developed algorithms in detail. Moreover, we used various versions of the dataset. We divided our experiments in two case studies:

    \begin{enumerate}[I.]
        \item First case study shows baseline results on the input dataset and explains benefits of the weighted approach with feature importances for drift detection.
        \item Based on baseline results at Case study I., we split the dataset and validate dataset stability improvements after the optimization. 
    \end{enumerate}

    Sec.~\ref{sec:discussion} summarizes achieved results and findings during all our experiments.
    

    \subsection{CS1: CESNET-TLS-Year22 Evaluation}
In the first case study we focus on the initial dataset stability benchmark for CESNET-TLS-Year22.
This experiment confirmed that guided retraining controlled by MFWDD vastly improved ML-trained classifier accuracy over time. The maintained model outperformed the initial model used as a baseline which is expected for long-term datasets. In the Fig.~\ref{fig:tls-baseline-retrained}, we may judge how the model reacted to dataset updates and compare it with benchmark for dataset without update in the Fig.~\ref{fig:tls-baseline-orig}. Vertical lines mark when the drift detection occurred, and the F1 score of the reference and the maintained models may be compared to illustrate model accuracy over time.


    \begin{figure*}
      \centering
      
    \begin{subfigure}{0.48\textwidth}
        \includegraphics[width=\linewidth]{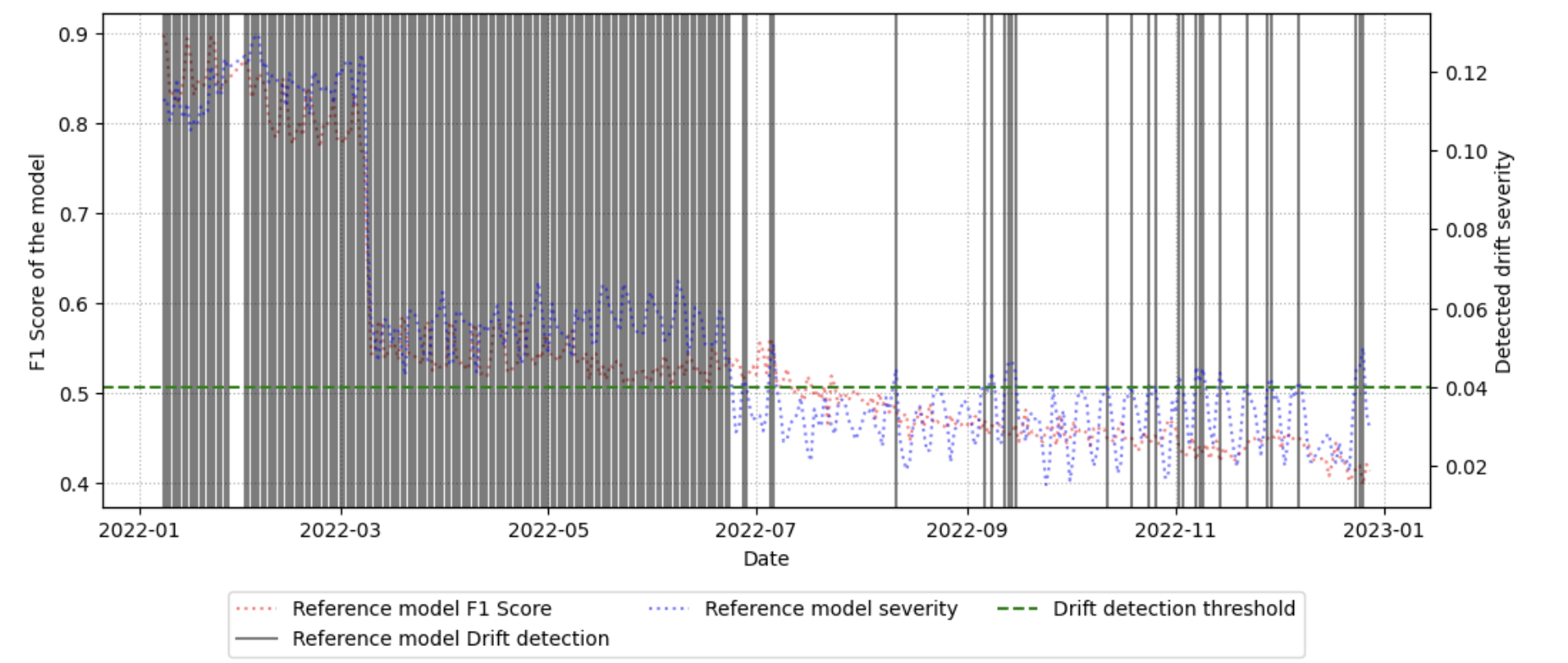}
      \caption{No retraining workflow}
      \label{fig:tls-baseline-orig}
    \end{subfigure}
    \hfill
    \begin{subfigure}{0.48\textwidth}
        \includegraphics[width=\linewidth]{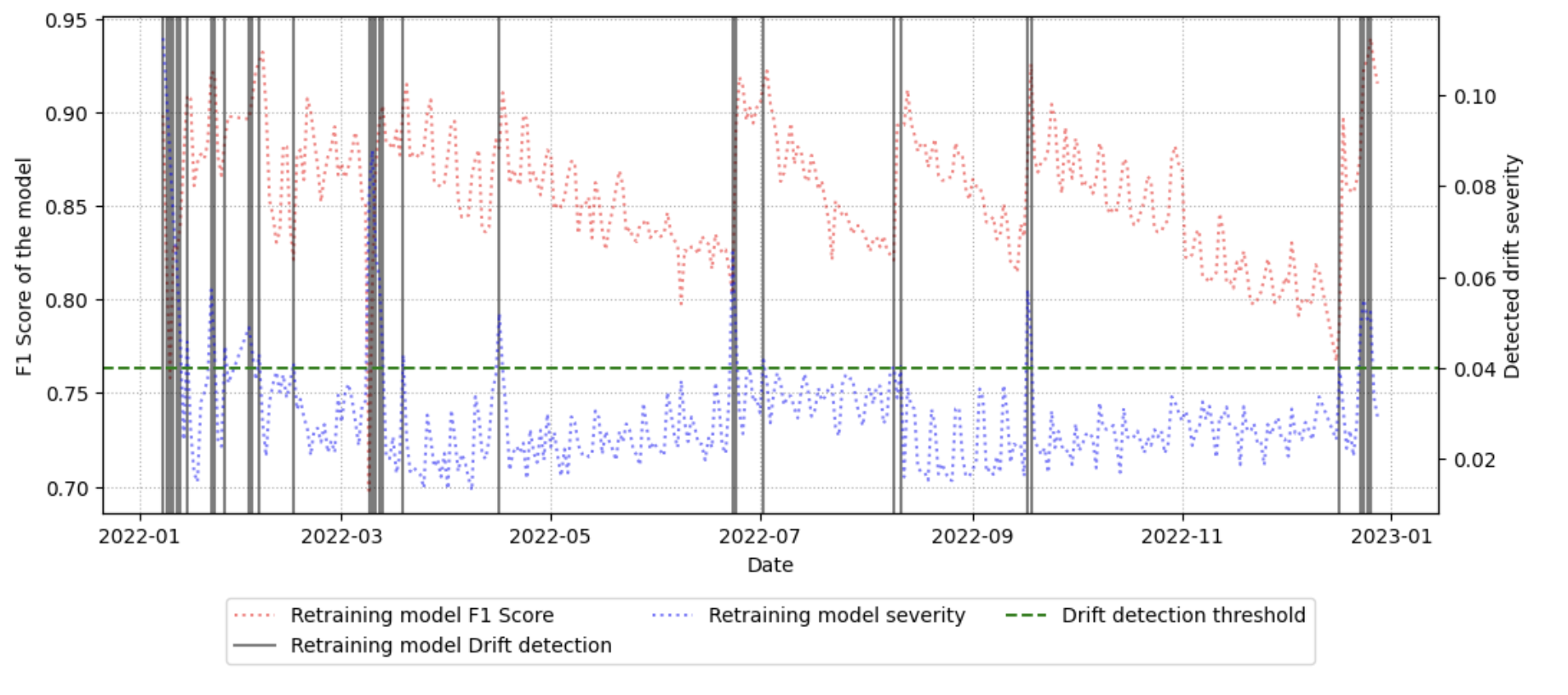}
        \caption{Retraining workflow}
      \label{fig:tls-baseline-retrained}
    \end{subfigure}
    \caption{Results of dataset stability benchmark for all classes in CESNET-TLS-Year22}
    \end{figure*}


As can be seen in Fig.~\ref{fig:tls-baseline-retrained}, in some cases, a single retraining was sufficient to keep the model precise for several months. In other cases, several retrainings were needed to improve the model's performance significantly. This may be due to the incremental drift being present, where the change of concepts may be gradual. Note that these results confirm the importance of well-timidness for model retraining and its impact on the AL efficiency. A single retraining can only lead to the learning of the intermediate concept, and subsequent retrainings might be needed while the new concept stabilizes. The plot of reference drift severity in time helps us understand drift severity if the model was not updated. The periodic behavior~\cite{janvcivcka2024analysis} is also seen each weekend and sometimes the drift detection threshold is almost attained. Nevertheless, the weekends do not have a significant impact on the model performance, which is in accordance with Jančička et al.~\cite{drift-janciluk}. 

After the initial ML model and dataset stabilization during the first weeks, a significant drop in the F1 score and a peak in the drift strength occurred in March. This also makes the initial version of the dataset unusable. The MFWDD method thus triggered several consecutive retrainings to adapt to new network conditions. The drift is attributed to changes in the network monitoring infrastructure and has impacted the majority of classes. Then, we can see recurring drifts every few months, that are mainly driven by \textit{SIZE\_1}, \textit{SIZE\_4} and \textit{DIR\_4} features representing \nth{1} and \nth{4} packet in the network flow. This is expected behavior for the network traffic dataset with public services since there are regular changes of certificates that are exchanged during first packets as confirmed by Luxemburk et al.~\cite{luxemburk2023encrypted}. At the end of the year, an incremental drift occurred and resulted in a series of retrainings to improve and stabilize the model performance. 

The average drift strength and amount of feature drift differ for both strategies: the updated model has a variance ranging from 0\,\% to 15\,\%, whereas with the no-retrain strategy, drifted features variance ranges from 1\,\% to 40\,\%.

For the retraining strategy, we can also notice a decrease in the F1 score between drift detections (e.g., the period between May and July) that happened repeatedly multiple times during the year. To achieve a higher F1 score, we can optimize MFWDD detection thresholds that will trigger ML retraining more often. However, the main focus of this work is to provide dataset benchmark to asses stability over time. Based on these results, we can see that the dataset complexity is high because F1 score is going down without significant impact on the drift strength score.
\begin{figure*}
     \centering
     \vspace{-1em}
     \includegraphics[width=0.6\textwidth]{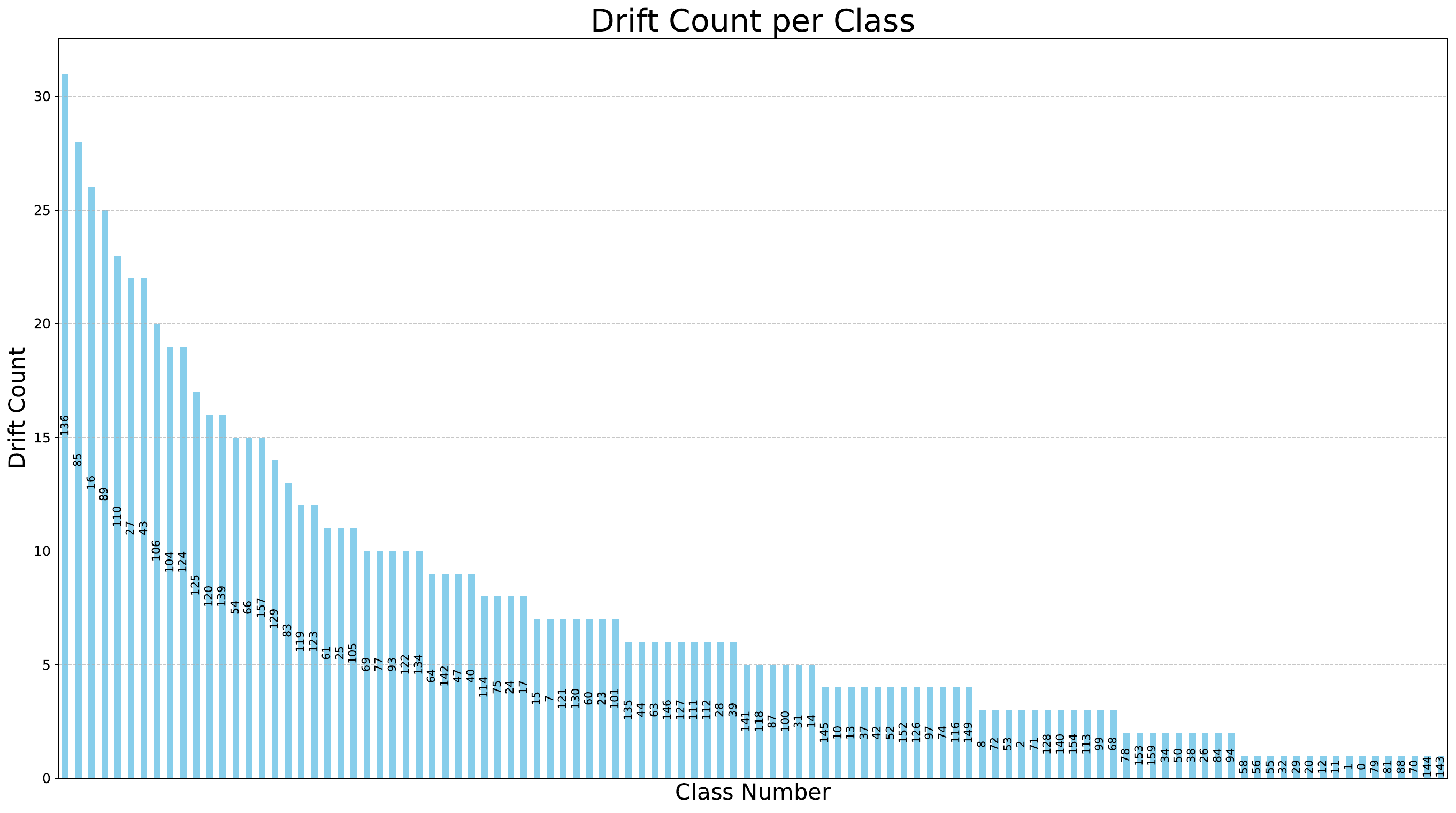}
     \caption{Number of drifts per class. Note that 54 classes did not drift at all and are not plotted to save space.}
       \label{fig:orig-class-drift}
\end{figure*}

From the class level results, shown on Figure~\ref{fig:orig-class-drift}, we can see that 83 out of 159 classes are drifted at least once and the number of detected drift events is not equally distributed. This shows that several classes make a bigger contribution to the global drift detection. With these insights, we can see the complexity of the CESNET-TLS-Year22 dataset that causes F1 score drops in the dataset baseline. Therefore, in the next case study Sec.~\ref{sec:cs2}, we validated results with split based on problematic and non-problematic classes. 

During the initial evaluation of the CESNET-TLS-Year22 dataset, we also compared the results of the MFWDD method when the weighting approach using feature importance is used and not used. Achieved outputs are summarized in~Tab.~\ref{tab:mfwdd-compare}.
Based on the summarized results, we can see that detection sensitivity for non-retraining workflow is significantly different. As we mentioned above, the ML classifier is more dependent on features representing the initial packet in the communication. Therefore, drift strengths more correlate with F1 score and identify events for these features more precisely which is important for later dataset analysis.
For CESNET-TLS-Year22, we have data drift almost immediately at the beginning, therefore, we can see drift strength sensitivity in a lower number of detections. However, for different datasets we can miss retraining trigger since we don't consider role of feature importances. 
This can be partially visible for experiments with the retraining workflow. However, due to high dataset complexity, the is no significant improvement in F1 score. 

To summarize, even without feature importance weights, we can see similar results, but they are not optimal. Especially for long feature vectors, there is lower sensitivity in the drift strength.

\begin{table*}[ht!]
\centering
\caption{Drift detection results}
\begin{tabular}{|l|r|r|r|r|}
\hline
 & \multicolumn{1}{l|}{\textbf{Drift detection count}} & \multicolumn{1}{l|}{\textbf{Drift strength mean}} & \multicolumn{1}{l|}{\textbf{Share drifted features mean}} & \multicolumn{1}{l|}{\textbf{F1 mean}} \\ \hline
Weights, no retraining & 191 & 0.053091 & 0.142617 & 0.546458 \\ \hline
Weights, retraining & 33 & 0.029130 & 0.032036 & 0.857492 \\ \hline
No weights, no retraining & 335 & 0.085893 & 0.296552 & 0.546458 \\ \hline
No weights, retraining & 28 & 0.025683 & 0.017008 & 0.855542 \\ \hline
\end{tabular}
\label{tab:mfwdd-compare}
\end{table*}

\subsection{CS2: CESNET-TLS-Year22 Class Separation}\label{sec:cs2}
In this case study, we follow up on results and observations from the previous case study. Since many classes are not drifted or drifted very rarely, it is not necessary to replace them with new data anytime the global drift is detected. Based on this various behavior of classes, we decided to split the dataset and evaluated that dataset stability benchmark. The initial benchmark result in Fig.~\ref{fig:tls-baseline-retrained} identifies 10 main data drift events. Therefore, we took all classes with more than 9 drifts (28 classes in total) for one subset of CESNET-TLS-Year22 dataset and the rest of classes (163 in total) were part of the second subset. 

With the most drifted classes removed, we can see benchmark results for reference workflow in Fig.~\ref{fig:split-non-problematic-referenced}. As expected, there is no significant improvement, and after the major drift event in March the ML classification is inaccurate. The results of the retraining workflow is depicted in Fig.~\ref{fig:split-non-problematic-retrained}. We can see that the F1 score is more stable during the whole year. The overall mean of the F1 score is 91\%, which is an improvement of 5\% from the initial dataset with all classes. Moreover, we identified 53 drift events that caused retraining. This is 20 more detected drifts in comparison with the original dataset.
Nevertheless, the complexity is reduced since there are no significant drops in the F1 score. Regarding the list of most problematic features, we can see a similar trend as for the original dataset. The most drifted features represent the initial packet in the network flow communication (\textit{DIR\_4}, \textit{SIZE\_4}, \textit{SIZE\_3}).

Next, we can see benchmark results for the group of 28 classes of problematic classes. With the referenced workflow depicted in Fig.~\ref{fig:split-problematic-referenced}, we don't see any unexpected trends, and the report is in line with previous variants where the major drift in March negatively affected the rest of the network traffic classification. We can notice that the benchmark visualization contains more white spots without a vertical line. This represents days where no traffic was received for the selected classes. The retraining workflow is depicted in Fig.~\ref{fig:split-problematic-retrained}. In general, we can see that selected classes are less stable and predictable since there is a larger standard deviation for the F1 score and deviation strength.

In summary, we created two variants of CESNET-TLS-Year22 based on class drift detections. With the proposed benchmark methodology, we evaluated created dataset variants and proved the benefits of this approach. Especially for large and long-term datasets, it is important to consider dataset maintenance to ensure the performance is continuously good enough.

  \begin{figure*}
      \centering
    \begin{subfigure}{0.48\textwidth}
        \includegraphics[width=\linewidth]{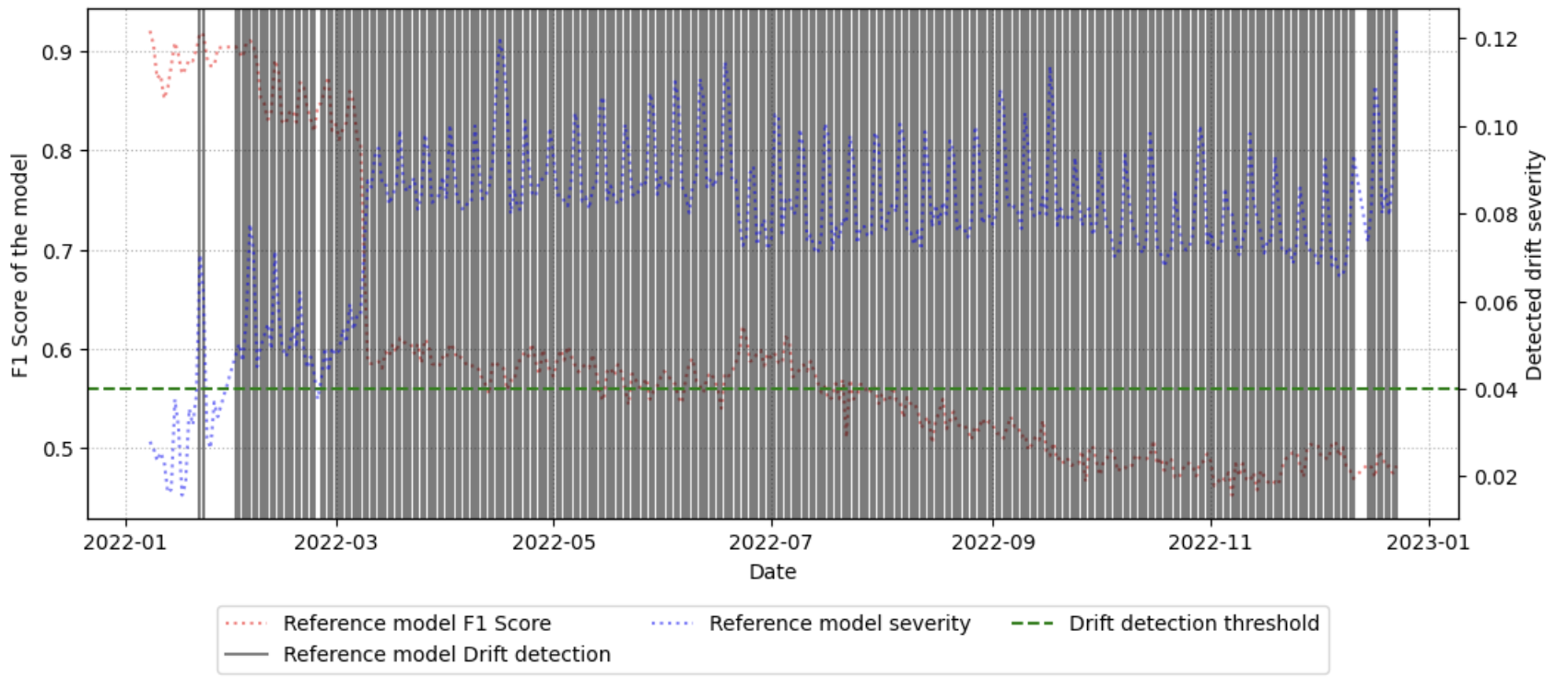}
      \caption{No retraining workflow}
      \label{fig:split-non-problematic-referenced}
    \end{subfigure}
    \hfill
    \begin{subfigure}{0.48\textwidth}
        \includegraphics[width=\linewidth]{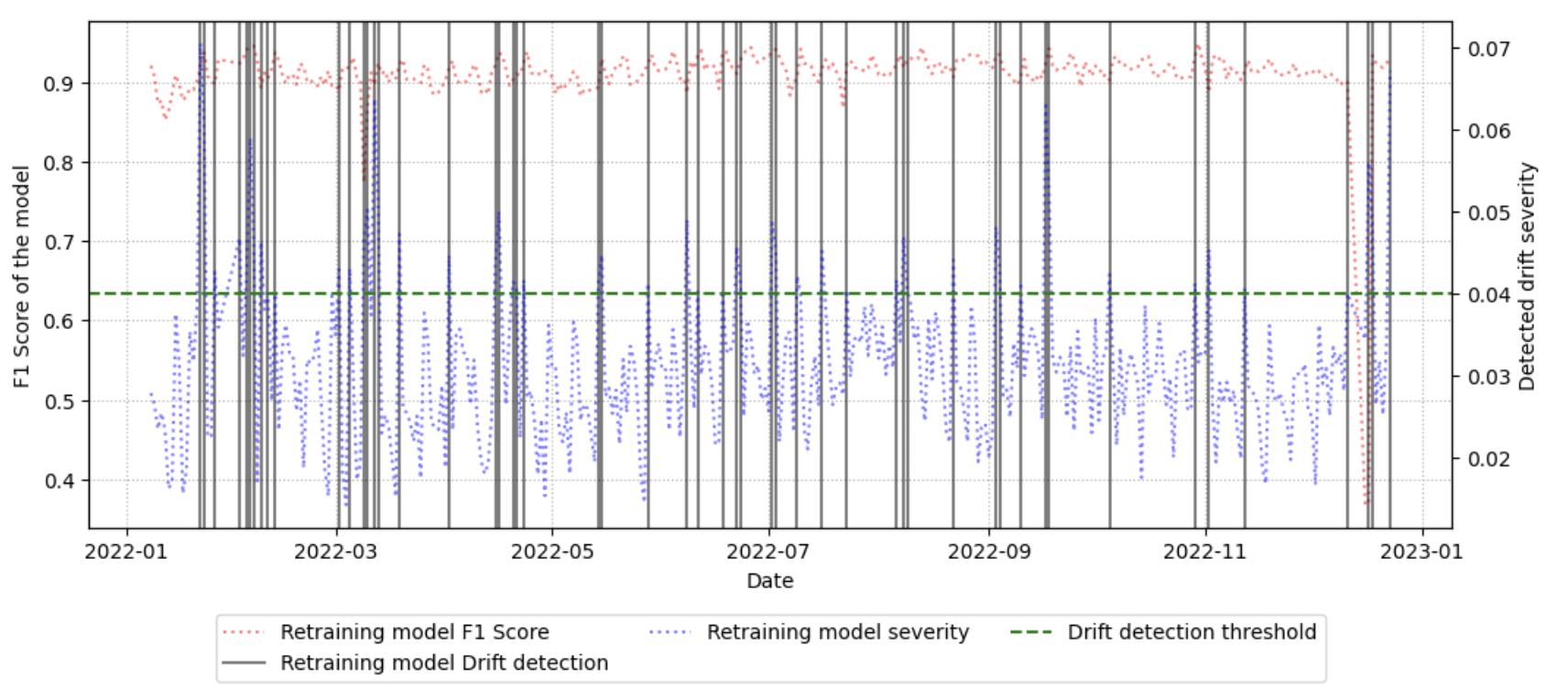}
        \caption{Retraining workflow}
      \label{fig:split-non-problematic-retrained}
    \end{subfigure}
    \caption{Results of dataset stability benchmark for non-drifted classes from CESNET-TLS-Year22 based on \ref{sec:cs2} selection}
    \end{figure*}

  \begin{figure*}
      \centering
    \begin{subfigure}{0.48\textwidth}
        \includegraphics[width=\linewidth]{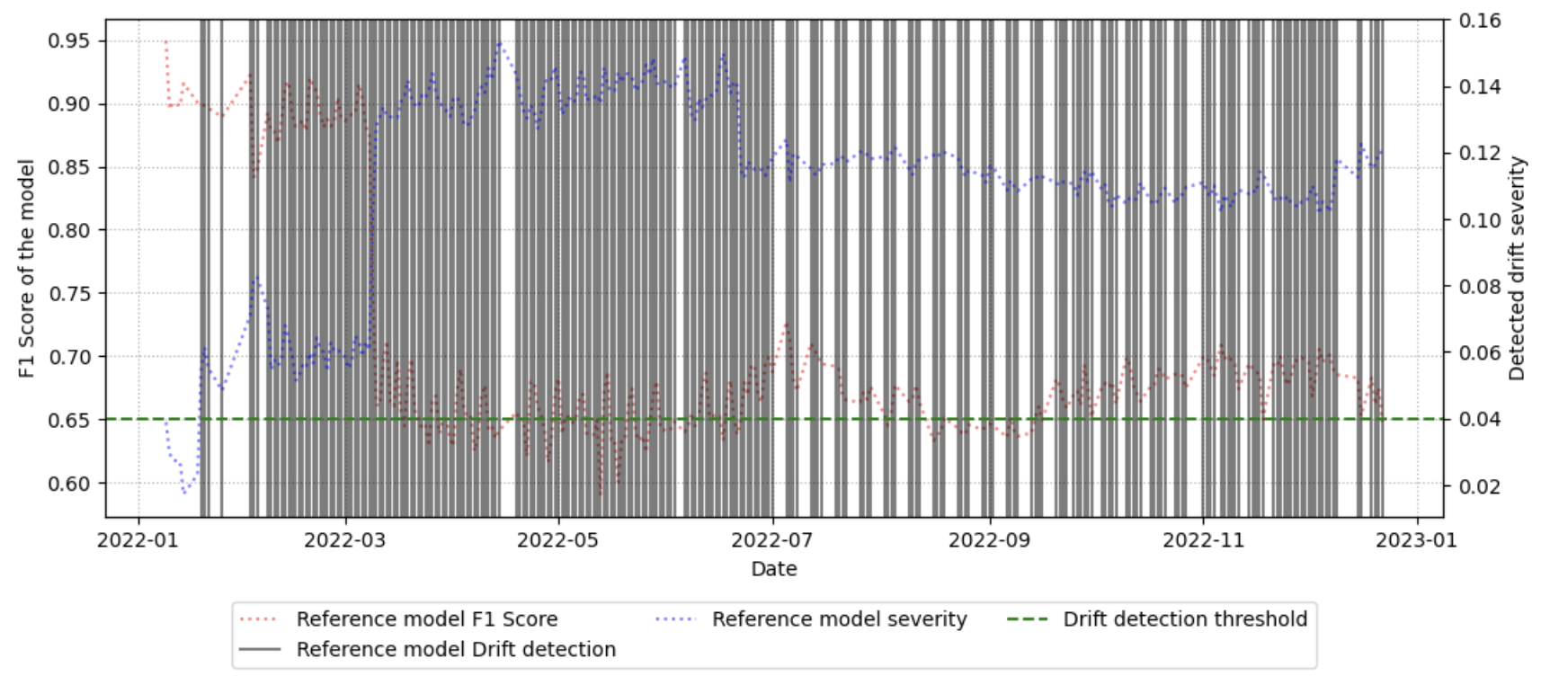}
      \caption{No retraining workflow}
      \label{fig:split-problematic-referenced}
    \end{subfigure}
    \hfill
    \begin{subfigure}{0.48\textwidth}
        \includegraphics[width=\linewidth]{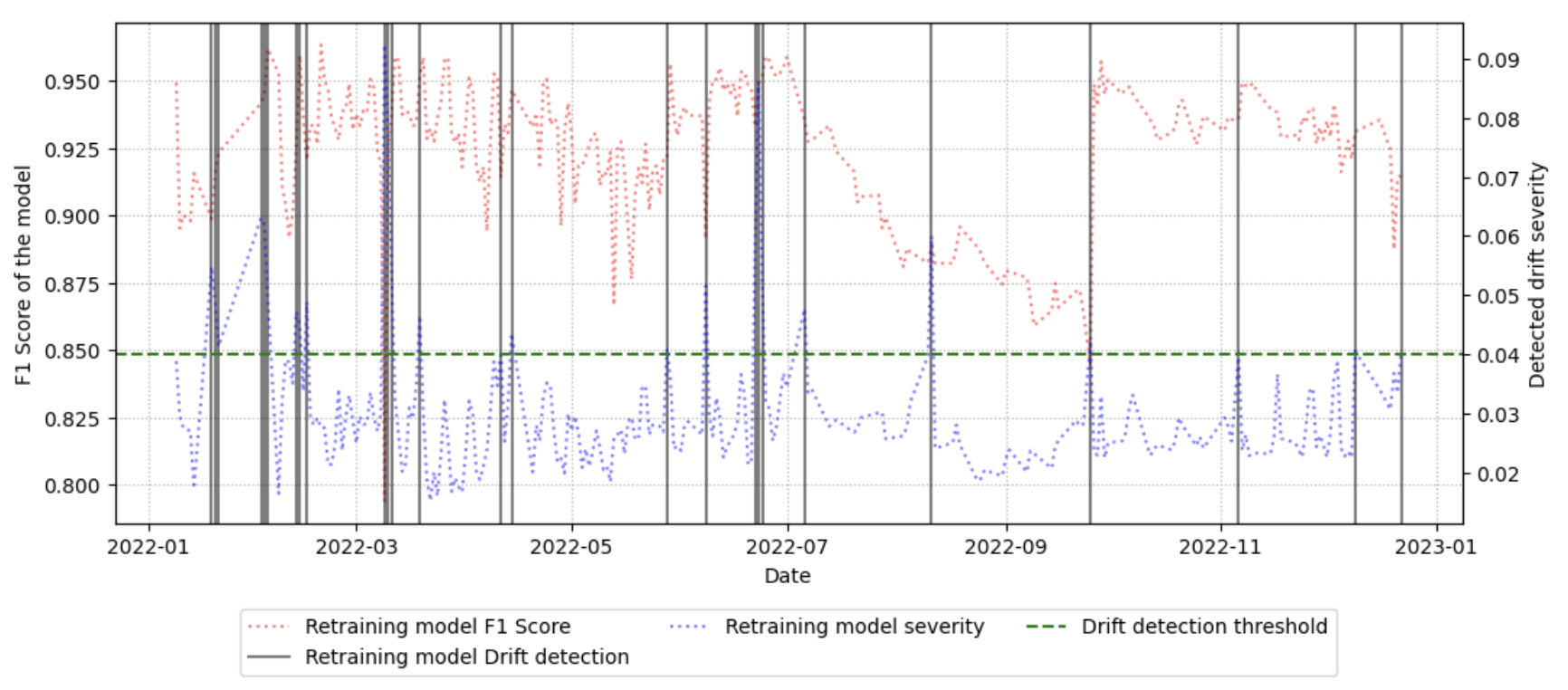}
        \caption{Retraining workflow}
      \label{fig:split-problematic-retrained}
    \end{subfigure}
    \caption{Results of dataset stability benchmark for top drifted classes from CESNET-TLS-Year22 based on \ref{sec:cs2} selection}
    \end{figure*}





\section{Discussion and best practices} \label{sec:discussion}

In this section, we briefly summarize leading outcomes and recommendations for best practices that stem from the experimental evaluation of the proposed dataset stability benchmark and analysis of data drifts and their detection by the MFWDD detector. The major findings and recommendations are described below.

It is always desirable to evaluate each feature drift severity before using it in the final model. If a feature that exhibits frequent drifts is present in the dataset, it will cause a disfavored performance decrease over time, and in extreme cases, its presence can lead to excessive model retraining. On the other hand, it is important to note that it is often the case that features exhibiting drifts encompass internal specifics for certain network traffic classes and are thus significant for the dataset. The offline setting of our experiments enabled us to evaluate the stability of each feature using the MFWDD together with the benchmark workflow. 

Our experiments imply that for network traffic classification problems, models based on the SPLT are affected by a large number of detected data drifts throughout long time periods, for example, a year. Therefore, the basic XGBoost representative with SPLT as input features is not so stable. Moreover, from related works by Malekghaini et al. \cite{malekghaini2022data,malekghaini2023deep}, we knew that even highly optimized Deep Learning models are not stable in time. In the related works, this problem was handled using complete periodic retraining, which cannot be easily done in all classification problems. Therefore, we suggest concentrating not only on the highest precision of trained models but also on their long-term stability because today's models significantly lose precision in a week or two.

Even though periodic retraining is the most straightforward solution in some domains, it brings technical and theoretical questions, such as: How do we gather and manage training datasets? How to detect data obsolescence and how to filter undeserved records. What records should be added to a dataset, on the other hand?  Generally, even if it is possible to use periodic retraining, we should avoid it for several reasons: 
\begin{enumerate}[I.]
    \item Most of the retraining will be unnecessary because no changes in data appear.
    \item Frequent or periodic, ML model retraining is highly resource-intensive. Furthermore, for complex problems, this approach might be unmanageable or unfeasible~\cite{9671779}.
    \item Frequent retraining without significant changes in data can cause fitting of the model on non-stable features because the model becomes too tailored to specific subsets of data, which often change to gain better performance by an insignificant percentage, reducing its generality \cite{chen2024lara}.
    \item It neglects some real-world effects, such as the weekend phenomenon described in \cite{janvcivcka2024analysis} or other recurring events.
\end{enumerate}

In the identified case studies we also showed that it is important to validate class separation, especially in scenarios with long list of class categories. The proposed benchmark workflow is capable to provide insights on the class level. Therefore, it can identify class similarities or differences to group sub-datasets. 
There are several existing papers  that can calculate flow similarity metrics~\cite{wang2014effective, bushart2020padding} and identify different dataset separation. During our experiments we tested similar approach where we used k-medoids with several popular metrics in the related work. Firstly, we identified that CESNET-TLS-Year22 contains a lot of user traffic which has similar SPLT pattern and k-medoids selected most of the traffic flows to the top five clusters. Secondly, we tried to encode the whole dataset using k-medoids clusters to validate quality of label assignments. To achieve this, we train k-medoids on the first week and run k-medoids prediction on the rest of the dataset. Even though, we got high F1 score for initial days, the overall performance was very low. Since the k-medoids were trained on the first week only, once data drift happened the k-medoids based encoding was not relevant. We tested this approach for various number of clusters and the result was always the same. This is one of use cases how the proposed benchmark stability workflow can help with validation of dataset modification questions.
Even though this initial idea of class separation was not optimal, we are missing a class similarity method that would be validated on large and complex datasets.

\section{Future work}\label{sec:future}


    The developed MFWDD framework provides promising methods for network drift detection and, thanks to its extensibility, enables an additional use of statistical methods. In future work, we would like to focus on deeper analysis of datasets
    and automatically identify drift type together with automatic insights with suggestions for dataset optimization.
    In related works, researchers observed several categories of drift patterns, sudden, gradual, incremental, and recurrent, Gama et al.~\cite{drift_survey}, and Lu et al.~\cite{datadrift}, but more research on drift types and their impact to a dataset of ML model quality is needed. This will reduce the process of dataset quality validation and optimization. Especially due to the complexity of collecting network traffic datasets, there is a demand for proactive analysis to avoid errors that can lead to fake results~\cite{nids-errors,lanvin-errors-CICIDS2017}. 

    Our future work also includes a development of an online catalogue of publicly available datasets for the networking domain. There are many public datasets, however, to our best knowledge none of them is providing unified structured description with validation metric. Therefore, we published the proposed benchmarking workflow with MFWDD method as online website\footnote{\url{https://dataset-catalog.liberouter.org}}. Except benchmarking output, there is unified and structured description of the dataset. The description contains generic information about the dataset content, research papers analysis, collection workflow and author, but it also contains qualitative metric to assess suitability for the demanded use case. We plan to continue in development of this dataset catalogue and add defined description for more existing dataset. To speed up and simplify the process of dataset description we plan to leverage AI agents that can run automatically individual tasks and combine outputs to the description structure.


\section{Conclusions} \label{sec:conclusion}
    We introduced a novel method called Model-based Feature Weight Drift Detection (MFWDD) together with benchmarking dataset stability workflow. 
    The main aim is to get a benchmarking report of dataset stability that identifies weak points and allows validation of the impact of changes. The benchmarking workflow is based on drift detection, and it is designed even for large datasets. 

    The proposed solution was evaluated using the popular and long-term public dataset CESNET-TLS-Year22. We explained the initial results of the benchmarking workflow, which were used for further optimization. Thanks to the detailed insights, we noticed drift differences among classes. After the dataset was split based on the class behavior, we got more stable and predictable results. 

    There are many options for how to use this framework. It works well in unsupervised or supervised scenarios. Therefore, there is no strict need for labels to compare dataset stability behavior on the global dataset level. That means we can perform detection on problems for which it is very difficult/expensive to get labels (for example, by using active scanning or performing human annotation).
    The proposed framework can be used in offline mode to provide dataset report and stability validation but also in online mode (e.g., part of Active Learning). In online mode, we can continuously use defined metrics to monitor selected dataset and provide retraining recommendations. This avoids unnecessary retrainings and improves dataset maintenance. 

    We made the proposed solution (MFWDD framework and dataset stability benchmark), containing both supervised and unsupervised methods, publicly available as an open-source project in the GitHub repository. The source code and examples are published for possible replication of the described results, for evaluation of additional datasets, or for follow-up on this research.

\bibliographystyle{IEEEtran}
\balance
\bibliography{main}

\end{document}